\documentclass{article}

\usepackage[english]{babel}
\usepackage{amssymb}
\usepackage{url}
\usepackage{graphicx}
\usepackage{multirow}
\usepackage{caption}
\usepackage{subcaption}
\usepackage[letterpaper,top=2cm,bottom=2cm,left=3cm,right=3cm,marginparwidth=1.75cm]{geometry}

\usepackage{amsmath}
\usepackage[colorlinks=true, allcolors=blue]{hyperref}

\title{OICycle: Observe, Infer, and Intervene Cycle for Belief-State Learning in Longitudinal Clinical Decision Making}

\author{Jitindra Fartiyal, Pedro Freire, Sergei K. Turitsyn \& Sergei G. Solovski\\
Aston Institute of Photonics Technology\\
Aston University\\
Aston Street, B4 7ET, Birmingham, UK\\
\texttt{240304225@aston.ac.uk, p.freiredecarvalhosourza@aston.ac.uk, } \\
\texttt{s.k.turitsyn@aston.ac.uk, s.sokolovsky@aston.ac.uk}\\
}

\begin{document}

\newcommand{\ARfull}{0.347}     % Task 1 AR, macro-averaged over 4 categories
\newcommand{\ARsingle}{0.253}
\newcommand{\CRfull}{0.601}     % Task 1 NR@10, macro-averaged over 4 categories
\newcommand{\CRsingle}{0.446}
\newcommand{\TtwoFull}{0.351}   % Task 2 macro-F1
\newcommand{\TtwoSingle}{0.293}
\newcommand{\TthreeFull}{0.649} % Task 3 macro-F1
\newcommand{\TthreeSingle}{0.585}
\newcommand{\nPatients}{72{,}075}

\maketitle

\begin{abstract}
\label{abstract}
Clinical care is a closed loop: clinicians observe partial evidence, infer the patient's state, and intervene, and each intervention determines what is observed next. This makes clinical decision making a partially observable Markov decision process (POMDP), in which a belief over a latent patient state must be updated from irregular, multimodal, and incomplete records. Yet most deep clinical models predict from a single snapshot or from coarsely aggregated history, while POMDP-based clinical methods need hand-specified discrete state spaces and transition models that real electronic health records cannot supply. We introduce \textbf{OICycle} (Observe, Infer, and Intervene Cycle), which learns a stochastic belief state directly from retrospective trajectories. At every episode, a learned recurrent operator fuses a sample of the previous belief with the new multimodal observations and the intervention that produced them. The resulting Gaussian belief is trained to predict clinicians' next interventions, the time to the next episode, and next-episode mortality. To enable reproducible evaluation, we release a benchmark derived from MIMIC-IV, MIMIC-IV-Note, and MIMIC-CXR that organizes \nPatients{} hospital admissions (5.4M laboratory, microbiology, radiology, prescription, and note records) into temporally ordered therapeutic--diagnostic episodes. Compared with an otherwise identical history-free model, OICycle improves next-intervention adaptive recall from \ARsingle{} to \ARfull{}, time-to-next-episode macro-F1 from \TtwoSingle{} to \TtwoFull{}, and next-episode mortality macro-F1 from \TthreeSingle{} to \TthreeFull{}. Analysis shows that the learned belief encodes intervention type and trajectory position and evolves smoothly across episodes, and that the benefit of history is concentrated in interventions whose ordering depends on prior findings.
\end{abstract}

\section{Introduction}
\label{sec:introduction}
\begin{figure}[t]
    \centering
    \includegraphics[width=\linewidth]{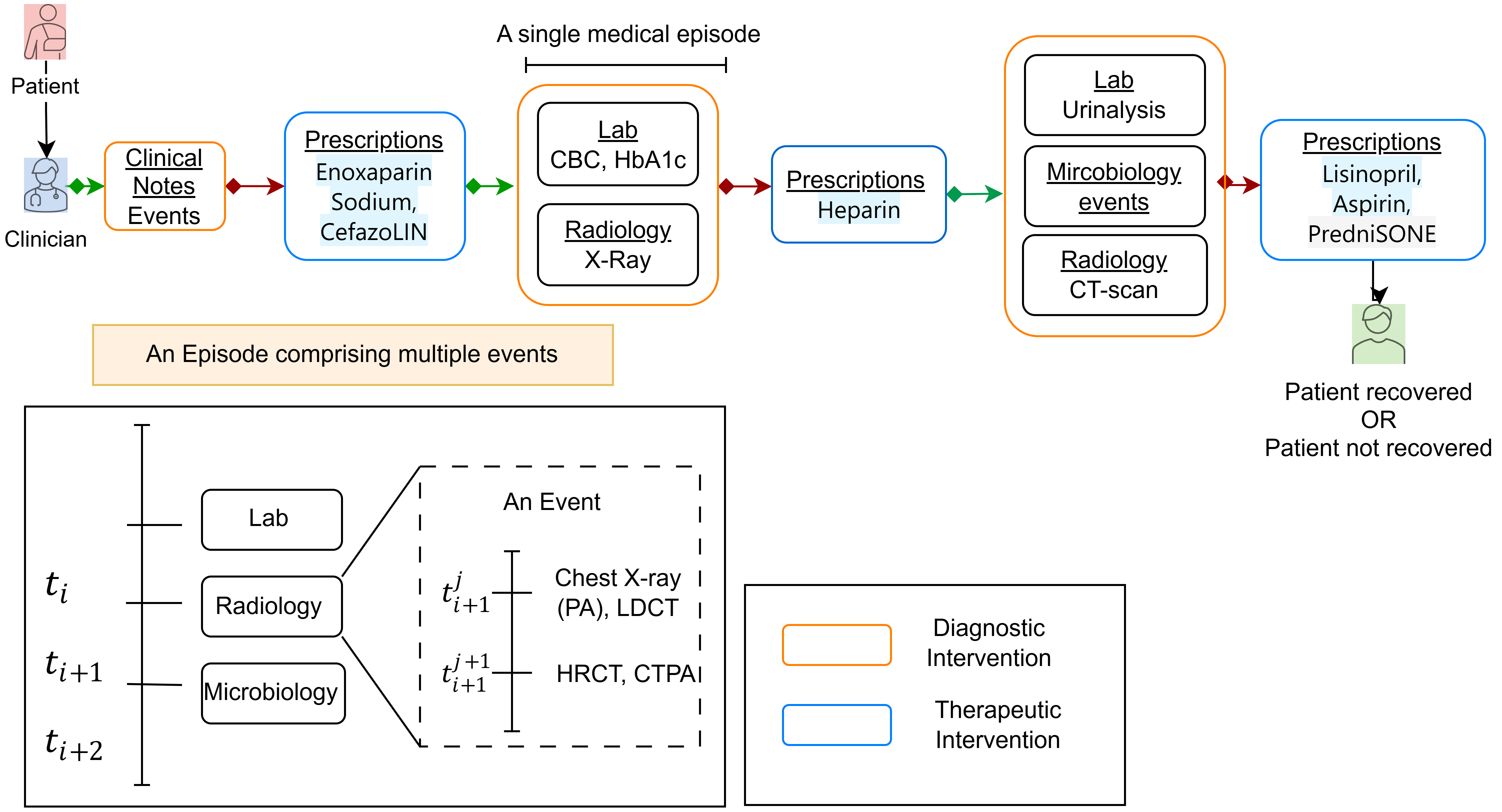}
    \caption{The clinical decision-making cycle often involves a follow-up diagnostic-therapeutic intervention cycle of episodes. Each episode can have multiple events, and each event can have multiple records.}
    \label{fig:patient_medical_trajectory}
\end{figure}
Longitudinal Clinical Decision Making (LDCM) is a sequential process in which clinicians act as a patient’s state evolves, spanning single admissions to years of care.  Figure \ref{fig:patient_medical_trajectory} presents this process.  The history of interventions is crucial but often overlooked by AI models that consider only isolated snapshots, leading to suboptimal decisions. Clinical data are also heterogeneous, irregular, incomplete, and only partially observe the true patient state, further complicating LDCM (\cite{jayaraman_primer_2024, najafi_advancements_2026, yu_reinforcement_2021}).

Prior work on POMDP-inspired frameworks has largely focused on specific diagnostic or treatment tasks. It does not comprehensively address the broader complexity of real-world clinical trajectories, which are diverse and irregular, and often rely on regular, homogeneous data and limited intervention types. As a result, these methods do not fully capture the fine-grained temporal and multimodal complexity of real longitudinal patient care.

\textbf{OICycle: Observe, Infer, and Intervene Cycle} addresses this gap by introducing a unified, predictive belief-state learning framework that models the temporal evolution of patient trajectories through recurrent therapeutic-diagnostic cycles. Existing latent-dynamics models often learn latent representations by modeling or reconstructing observed clinical data. This objective can be limiting for clinical data for two reasons. First, observations span free text, images, and thousands of sparse codes, so reconstruction spends capacity on detail that does not matter for decisions. Second, an observation exists only if a clinician chose to acquire it, so its absence is itself informative and an observation likelihood would have to model the ordering policy too. OICycle departs from these models in three ways that are, to our knowledge, not previously combined:
\begin{itemize}
    \item It reformulates the clinical POMDP as predictive belief-state learning: the belief is shaped by its ability to forecast future clinical events rather than regenerate observations, removing the need for an explicit state space, transition model, or observation model.
    \item It makes the \emph{action-conditioned therapeutic-diagnostic episode} the unit of belief update, so that every update is tied to the intervention that caused the new evidence. Sequence models that treat interventions as covariates, and clinical POMDPs defined over fixed time steps, both lack this structure
    \item It pairs the model with a benchmark that, unlike existing MIMIC-derived resources, exposes this action-observation structure across modalities, making such models evaluable. 
\end{itemize}
 In summary, this paper makes the following key contributions:
 \begin{itemize}
    \item We introduce OICycle as a sequential predictive latent belief-state learning framework for longitudinal clinical decision-making.
    \item We release the data preprocessing code to provide open access to a derived MIMIC dataset that organizes patients’ clinical trajectories into meaningful clinical episodes  
    \item We evaluate our results on three predictive tasks: next-intervention selection, time-to-next-episode forecasting, and next-episode mortality prediction, compare them against a single-snapshot clinical model, and analyze the predictive belief-state representation.
\end{itemize}

\section{Related Works}
\label{sec:related_works}
Early works (\cite{li_behrt_2020, lee_biobert_2020, huang_clinicalbert_2020, choi_gram_2017, rasmy_med-bert_2021, peng_transfer_2019}) primarily learn local clinical context using BERT-based models (\cite{devlin_bert_2019}) or use a graph model to structure relationships between medical concepts.  However, these approaches do not account for the patient’s medical trajectory.

Closest to OICycle are decision-making frameworks like DECIDE-Lab (\cite{allenbrand_decide-lab_2026}), which uses Value of Information for diagnostic selection and maintains a Bayesian posterior over explicitly defined clinical states. Similarly, DPWM (\cite{li_beyond_2026}) learns intervention-conditioned dynamics from recorded clinical trajectories to simulate counterfactual clinical scenarios, but focuses on preoperative risk assessment and intervention evaluation. By contrast, OICycle is designed for retrospective longitudinal data and learns a unified probabilistic predictive belief-state over a latent clinical representation that integrates observations from both historical interventions and diverse clinical events. While FORBES (\cite{chen_flow-based_2022}) also models a probabilistic belief state, it does so in a general problem framework, not in a clinical setting. It does not address the challenges present in clinical data. This distinction underlines OICycle’s focus on realistically structured longitudinal clinical trajectories.

(\cite{li_optimizing_2019, krishnan_deep_2015}) learn probabilistic belief states within POMDP/RL frameworks but operate at coarse temporal resolution (inter-episode). In contrast, OICycle models fine-grained, multimodal intra-episode data, which is prevalent in real clinical scenarios (\cite{tsoukalas_data_2015, futoma_popcorn_2020}), and updates Bayesian beliefs using discrete, manually defined states with estimated transition/observation probabilities. OICycle does not require a predefined discrete state space or explicit transition and observation probabilities; instead, it learns the belief representation directly from observed retrospective multimodal trajectories. DeepCare (\cite{pham_predicting_2017}) predicts outcomes using coarse temporal windows (multiple admissions), missing single-admission intra-episode complexities. The POMDP approach in (\cite{hauskrecht_planning_2000}) also relies on discrete, explicitly defined state spaces, unlike OICycle’s continuous latent modeling.

In summary, prior research has established key foundations, but challenges remain in jointly modeling fine-grained intra-episode temporal structure and heterogeneous modalities. Most methods do not fully leverage temporal dependencies between sequential interventions and subsequent observations. OICycle addresses these by unifying belief-state learning and predictive modeling across multimodal, asynchronous, longitudinal data.

Public longitudinal multimodal datasets are scarce and face challenges like high dimensionality, privacy, and trade-offs between resolution and storage (\cite{acosta_multimodal_2022}). MIMIC-IV (\cite{johnson_mimic-iv_2023}) is widely used but requires processing. HiRID-ICU-benchmark (\cite{yeche_hirid-icu-benchmark_nodate}) and EHRSHOT (\cite{wornow_ehrshot_nodate}) offer benchmarking pipelines but contain only ICU or EHR data. Various derived datasets exist for specific tasks, while preprocessing tools (\cite{yang_pyhealth_2023, wang_mimic-extract_2020}) enable standardized pipelines but are not tailored for fine-grained multimodal episode modeling.

Thus, while existing resources provide longitudinal and multimodal clinical data, OICycle's derived dataset organizes heterogeneous observations and interventions into clinically meaningful episodes, providing a standardized basis for benchmarking sequential clinical decision-making models.

\section{Preliminaries}
\label{sec:preliminaries}
\subsection{Problem Formulation: Clinical Decision Making Process}
\label{subsec:Problem_Formulation}
The clinical decision-making process can be formulated as a partially observable Markov Decision Process (POMDP) (\cite{kaelbling_planning_1998}). A POMDP is formally defined as a 6-tuple:
\begin{equation}
        <\mathcal{S, A, O}, \mathbb{T}, \mathcal{R}, \Omega> 
        \label{eq:pomdp_equations}
\end{equation}
\begin{equation*}
    \begin{aligned}
        &\text{where}, \mathcal{S, A, O} : \text{is a set of states, actions, and observations} \\
        &\mathbb{T}: \text{is the transition probability function } \mathbb{T}(s_{t+1}|s_t, a_t): \mathcal{S} \times \mathcal{S} \times \mathcal{A} \\
        &\mathcal{R}: \text{is the reward function } \mathcal{R}(s_t, a_t):  \mathcal{S} \times \mathcal{A} \\
        &\Omega: \text{is the observation probability function } \Omega(o_{t+1}|s_{t+1}, a_t): \mathcal{O} \times \mathcal{S} \times \mathcal{A} 
    \end{aligned}
\end{equation*}
However, real-world clinical data rarely provide the necessary information to derive these explicit transition and observation functions. Therefore, inspired by the POMDP framework, we abstracted the OICycle clinical decision-making process as sequential predictive belief-state learning, defined as a 4-tuple:
\begin{equation}
        <\mathcal{S, A, O, F_\theta}>
        \label{eq:oicycle_equations}
\end{equation}
where, $\mathcal{S, A, O}$ is a set of latent clinical patient states, interventions, and observations and $\mathcal{F_\theta}$ is the learned recurrent operator that updates the latent belief state from observations and interventions. At each prediction point, the model has access to the clinical history accumulated up to that point ($H_t$), defined as
\begin{equation}
    H_t = \{(a^*_0, o_1), (a^*_1, o_2) \ldots (a^*_{t-1}, o_t) \}
    \label{eq:history_accumulation_equation}
\end{equation}  
where, $a^*$ is the realized/ground truth clinical intervention(action), $a^*_0$ is the initial clinical assessment by the clinician, $a^*_{t-1}$ is the realized clinical intervention at episode $t-1$, and $o_t$ is the clinical observations at episode $t$.

\subsection{Clinical Episode Embedding}
At each episode, the model receives multimodal observations. These observations are acquired after the realized clinical action taken in the preceding episode and are therefore temporally coupled with that action. We construct the episode embedding $X_t$ as 
\begin{equation}
    X_t = E(o_t, a^*_{t-1}); \text{where, }E \text{ denotes the multimodal episode encoder.}
    \label{eq:episode_embedding_equation}
\end{equation}

\subsection{Learned Recursive Update Operator}
We then combined the sampled predictive belief from the previous episode with the embedding of the current clinical episode using a recurrent operator $F_\theta$. In our implementation, $F_\theta$ is instantiated as a Gated Recurrent Unit (GRU) (\cite{cho_learning_2014}), defined as
\begin{equation}
    r_t = F_\theta(z'_{t-1}, X_t)
    \label{eq:learned_recursive_update_operator}
\end{equation}
where, $r_t$ is the recurrent update representation produced by the operator $F_\theta$ at episode $t$ and $z'_{t-1}$ denotes a sample from the learned predictive belief distribution at episode $t-1$. 
    
\subsection{Probabilistic Belief-state parametrization}
We introduced a small belief network $g_\phi$ as a two-layer fully connected neural network that transforms the recurrent update representation into the parameters of learned predictive distribution over the latent belief-state representation. It is defined as 
\begin{equation}
    \begin{aligned}
    m_t = g_\phi(r_t); \mu_t = w_\mu m_t + b_\mu; \log \sigma^2_t = w_\sigma m_t + b_\sigma
    \end{aligned}
    \label{eq: belief_network}
\end{equation}
where $\mu,\log\sigma^2_t \in \mathbb{R}^{512}$ are mean and log variance of the learned predictive belief-state distribution.
We model the learned predictive belief-state distribution $q_\phi$ defined  as 
\begin{equation}
    \begin{aligned}
        q_\phi(z_t|H_t) \approx \mathcal{N}(\mu_t, diag(\sigma^2_t))
    \end{aligned}
    \label{eq:learned_belief_state_distribution}
\end{equation}
where, $z_t$ denotes the random variable for the learned latent predictive belief-state representation, $z_t \in\mathbb{R}^{512}$.

\subsection{Differentiable latent belief sampling}
To obtain a sampled latent belief representation $z'_t$ from the learned predictive belief-state distribution $q_\phi$ while preserving  differentiability, we use the reparameterization trick (\cite{kingma_auto-encoding_2022}) as defined as:
\begin{equation*}
    z'_t \sim q_\phi(z_t|H_t)
\end{equation*}
\begin{equation}
    \begin{aligned}
        &\text{Let } \epsilon_t \sim \mathcal{N}(0,I_{512}) \text{,   then } z'_t = \mu_t + \sigma_t \odot \epsilon_t 
    \end{aligned}
    \label{eq:belief_state_sampling}
\end{equation}
where $\sigma_t = \exp(\frac{1}{2}\log\sigma^2_t)$, $z_t$ is the latent predictive belief-state random variable at episode $t$, and $z'_t$ is the sampled latent belief representation at episode $t$.

\subsection{Latent belief-guided prediction}
Then the sampled latent belief representation is used to predict the next episode's clinical interventions (actions) as
\begin{equation}
    p_\psi(a_{t+1}|z'_t); a_{t+1}\in\mathcal{A}
    \label{eq:belief_guided_prediction}
\end{equation}
The prediction losses (defined in Eq. \ref{eq:task1_loss}) are computed against the corresponding realized clinical actions ($a^*_{t+1}$) and backpropagated through the sampled latent belief representation. The process is recursively applied across the patient's clinical episodes until the episode sequence terminates, resulting in a POMDP-inspired sequential predictive belief-state learning process OICycle.
\begin{equation}
    \label{eq:task1_loss}
    \mathcal{L}^t_{int} = \lambda_{ty} \mathcal{L}^t_{ty} + \lambda_{sty}\mathcal{L}^t_{sty}
\end{equation}
\begin{equation*}
\begin{aligned}    
    &\text{where, } \mathcal{L}_{sty} = -\sum_k w^{sty}_k[y^{sty}_{t+1, k}\text{ }log(\hat{y}^{sty}_{t+1, k}) + (1 - y^{sty}_{t+1, k}) log(1-\hat{y}^{sty}_{t+1, k})] \\
    &\mathcal{L}_{ty} = - \sum_k w^{ty}_k[y^{ty}_{t+1, k}\text{ }log(\hat{y}^{ty}_{t+1, k}) + (1 - y^{ty}_{t+1, k}) log(1-\hat{y}^{ty}_{t+1, k})] 
\end{aligned}
\end{equation*}
Here, $\mathcal{L}_{int}^t$ is the total loss, $\mathcal{L}_{ty}^t$ and $\mathcal{L}_{sty}^t$ are losses for intervention types and subtypes, $w^{ty}_k$ and $w^{sty}_k$ are class weights for type and subtype imbalance, and $\lambda_{ty}$, $\lambda_{sty}$ control the respective loss terms.

\section{Methodology}
\label{sec:Methodology}
\subsection{Clinical Dataset and Episode Construction}
\label{subsec:dataset}
We constructed a multimodal longitudinal episodic dataset by integrating MIMIC-IV \cite{johnson_mimic-iv_2023, pollard_physionet_2026}, MIMIC-IV-Note \cite{PhysioNet-mimic-iv-note-2.2, pollard_physionet_2026}, and MIMIC-CXR-JPG \cite{johnson_mimic-cxr-jpg_2019, pollard_physionet_2026}. From 364,627 patients in MIMIC-IV, we selected each patient’s first hospital admission with both a MIMIC-IV-Note and at least one DRG code, restricting to one admission per patient for consistency and preventing data leakage. The final cohort includes 72,075 patients with 349,489 unstructured text records, 5,057,180 structured clinical records, and 15,044 radiological images. Appendix \ref{subsec:dataset_plots} summarizes dataset statistics and episode distributions.

\subsubsection{Clinical Episode Construction}
\label{subsubsec:clinical_episode_construction}
A \textit{clinical intervention} is the smallest data unit, categorized as either \textit{Therapeutic Intervention} (TI; prescriptions) or \textit{Diagnostic Intervention} (DI; notes, labs, microbiology, radiology). Each may include multiple subtypes, such as several lab tests within a laboratory intervention.

A medical episode is a temporal unit containing either a TI or DI. TI and DI typically alternate but may occur consecutively; note-only DI episodes can appear at the start of a trajectory. Modality-specific temporal windows group interventions into episodes primarily by intervention type, and few conflicts were resolved when long gaps separate multiple single interventions by using a fixed temporal window per intervention type.

We define a \textit{TI-DI episode} as a temporally bounded clinical unit consisting of a TI followed by the subsequent DI, where the TI represents a realized therapeutic intervention and the DI represents the resulting clinical observations. In addition, organizing data into TI-DI episodes enables modeling of the sequential cause-and-effect relationships that are central to clinical reasoning. Consequently, this construct serves as the basic temporal unit for our sequential predictive belief-state learning framework.

\subsubsection{Prediction Tasks}
\label{subsubsec:prediction_task}
Our episodic representation supports three prediction tasks. First, the primary task predicts the intervention type and subtype of the next clinical episode, providing the main learning signal for belief-state representation. Second, we predict the elapsed time to the next episode by discretizing intervals into temporal bins, evaluating whether the representation captures expected timing. Only episodes with an observed subsequent episode are included. Third, we predict episode-level mortality in the subsequent episode. We exclude final episodes from episode-level mortality.

\subsection{Model Architecture}
\label{subsec:model_architecture}
\begin{figure}[t]
    \centering
    \includegraphics[width=\linewidth]{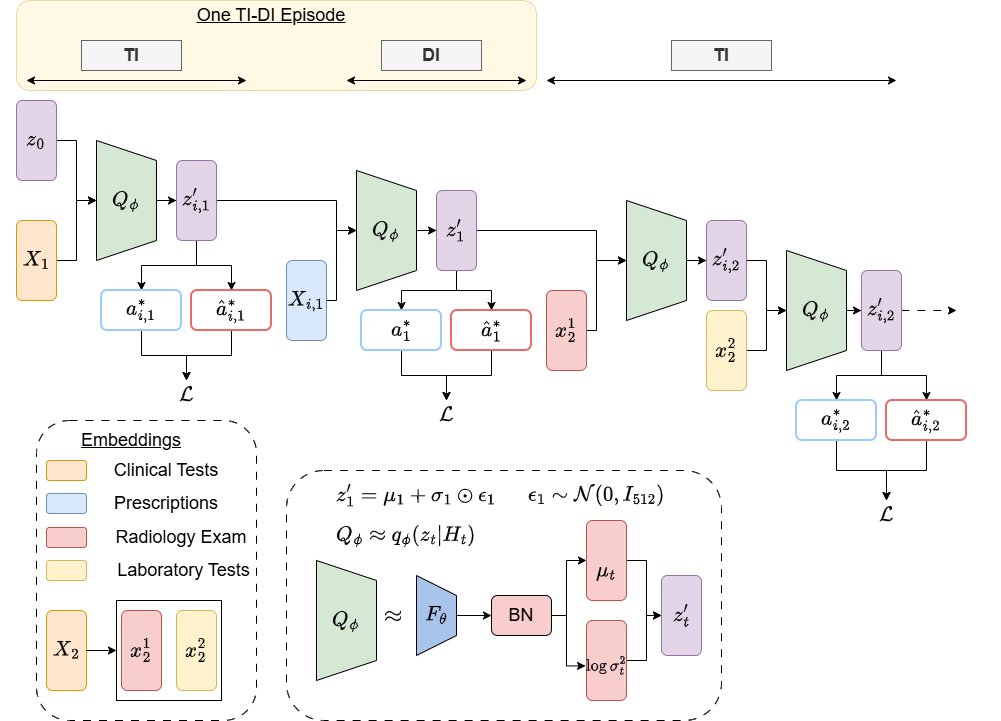}
    \caption{Overview of OICycle architecture. $z_0$ is the initial random state from where the sequential belief-state learning starts followed by a series of TI and DI episode.}
    \label{fig:overview_architecture}
\end{figure}

Figure \ref{fig:overview_architecture} presents an overview of the OICycle architecture. Initially, the model receives the initial state $z_0$ and clinical notes $X_1$, which are input into $Q_\phi$ (the learned predictive belief-state distribution). Within $Q_\phi$, $F_\theta$—the recurrent update operator implemented using a GRU—processes the inputs sequentially. The output of this sequence passes to a small belief network that produces the parameters of the learned predictive belief-state distribution. Next, an intermediate state $z'_{i,1}$ is sampled, where $i$ refers to the TI intervention within a TI-DI episode, facilitating the propagation of state information. At each prediction point, the model generates predictions and backpropagates the associated loss. If a single episode comprises multiple events (for example, $X_2$ includes a radiology exam embedding $x^1_2$ and a laboratory test embedding $x^2_2$), the system processes each event embedding in sequence and generates the prediction only after processing the final embedding. For medical text embeddings, used for both clinical notes and radiology notes, we use BioClinicalBERT \cite{ling_bioclinical_2023} to obtain the embeddings, followed by a projection layer to map them back to the common embedding dimension.  For some patients, radiology images are also available; in these cases, we use image embeddings instead of radiology note embeddings.

\section{Experiments \& Results}
\label{sec:experiments_results}
\subsection{Experimental Setup}
\label{subsec:Experimental_setup}
We ran two models: our proposed model, OICycle that learns from patient trajectory history, and OICycle-Single (OICycle-S), which predicts only from current-episode observations. Both share the same architecture, except that OICycle incorporates patient history. The initial state is a parametrized object initialized with Xavier uniform initialization, and the hyperparameters used are shown in the Appendix (\ref{sec:hyperparameters}). We ran these models for 10 epochs on an NVIDIA RTX 5000 Ada Generation GPU. We used our derived OICycle dataset for training, validation, and testing, split at a 70:15:15 ratio. The dataset used multilabel stratification on \textit{patient\_id}, \textit{drg\_code}, and task-related predictive targets. A data point in the dataset represents a single patient with all its episodic data.
As mentioned in (Section \ref{subsubsec:prediction_task}), we have three predictive tasks: Task 1, Task 2, and Task 3. For Task 1, intervention and sub-intervention type prediction, we used a class-imbalance-aware BCE loss defined in Eq. (\ref{eq:task1_loss}). 
For Task 2, the temporal prediction task, where the next-episode time interval belongs to one of $C$ mutually exclusive temporal bins, we use multiclass cross-entropy:
\begin{equation}
    \label{eq:task2_loss}
    \mathcal{L}^t_{time} = -\sum^C_{c=1} y^{time}_{t+1, c} log(\hat{y}^{time}_{t+1,c})
\end{equation}
And for Task 3: mortality prediction, we use binary cross-entropy:
\begin{equation}
    \label{eq:task3_loss}
    \mathcal{L}^t_{m} = - [y^{m}_{t+1} log(\hat{y}^{m}_{t+1}) + (1 - y^{m}_{t+1}) log(1-\hat{y}^{m}_{t+1})]
\end{equation}
Hence, the overall loss is then defined as: 
\begin{equation}
    \label{eq:overall_loss}
    \mathcal{L} = \sum_t[\lambda_{int}\mathcal{L}^t_{int} + \lambda_{time}\mathcal{L}^t_{time} + \lambda_{m}\mathcal{L}^t_{m}]
\end{equation}
where $\lambda_{int}$, $\lambda_{time}$, and $\lambda_m$ control the relative contributions of the three predictive tasks.

For Task 1 evaluation, because a clinical episode may contain multiple intervention subtypes and the number of relevant interventions varies across episodes, we use both adaptive and fixed-K retrieval metrics. Therefore, we introduce two derived metrics: \textit{AdaptiveRecall@K} (AR) and \textit{ClinicalRelevance@10} (CR). AdaptiveRecall@K measures the proportion of relevant interventions recovered within a prediction set whose size adapts to the number of relevant interventions in the ground truth of that particular episode. Whereas ClinicalRelevance@10 evaluates the clinical relevance of the top-10 predicted interventions. Their formulations are given in Eqs: \ref{eq:adaptiverecall}, \ref{eq:clinical_relevance}
\begin{equation}
    \mathrm{AdaptiveRecall@K} =
        \frac{1}{T} \sum_{t=1}^{T} 
        \frac{\left|\mathcal{Y}_t \cap \hat{\mathcal{Y}}_t^{K_t}\right|}
            {|\mathcal{Y}_t|},
        \qquad K_t = |\mathcal{Y}_t|.
    \label{eq:adaptiverecall}
\end{equation}
\begin{equation}
    \mathrm{ClinicalRelevance@10} =
        \frac{1}{T}    \sum_{t=1}^{T}
        \frac{
        \left|\hat{\mathcal{Y}}_t^{10}\cap\mathcal{Y}_t\right|
        }{
        \min\left(10,\left|\mathcal{Y}_t\right|\right)
        },
    \label{eq:clinical_relevance}
\end{equation}
Here, \(T\) is the number of evaluated episodes, \(\mathcal{Y}_t\) is the set of ground-truth interventions for the episode \(t\), and \(\hat{\mathcal{Y}}_t^{K_t}\) contains the model's top-\(K_t\) predictions, \(\hat{\mathcal{Y}}_t^{10}\) denotes the top-10 predicted interventions.
For Task 2 and Task 3, we use the macro-F1 score as the primary evaluation metric.

\begin{table}[t]
    \centering
    \caption{Results of OICycle and OICycle-S across all three predictive tasks. Values are reported with 95\% confidence intervals estimated using bootstrap resampling for statistical uncertainty assessment.}
    \label{tab:main_result_table}
    \begin{tabular}{c c c c c}
        \hline
        Algorithms & \multicolumn{2}{c}{Task 1} & Task 2 & Task 3\\
         & Mean AR & Mean CR & Mean F1 & Mean F1\\
         \hline
         OICycle & \textbf{0.347}(0.344-0.350) & \textbf{0.601}(0.597-0.604) & \textbf{0.351}(0.346-0.356) & \textbf{0.649}(0.622-0.674) \\
         OICycle-S & 0.253(0.252-0.255) & 0.446((0.443-0.449)) & 0.293(0.292-0.295) & 0.585(0.571-0.599) \\
         \hline
    \end{tabular}
\end{table}
\begin{figure}[t]
    \centering
    \begin{subfigure}[b]{0.49\textwidth}
        \centering
        \includegraphics[width=\linewidth]{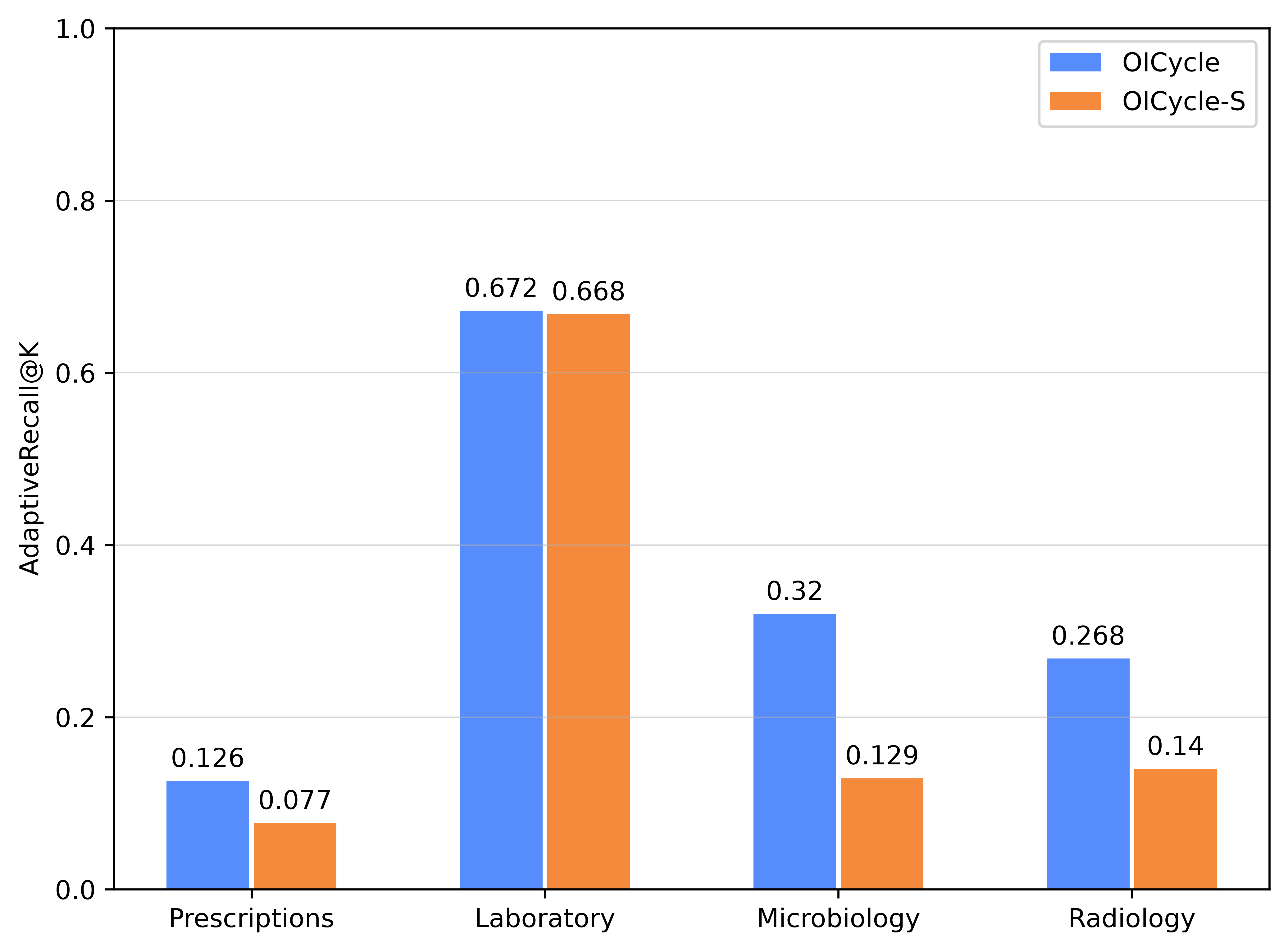}
        \caption{Adaptive Recall@K}
        \label{fig:intervention_ar}
    \end{subfigure}
    \hfill
    \begin{subfigure}[b]{0.49\textwidth}
        \centering
        \includegraphics[width=\linewidth]{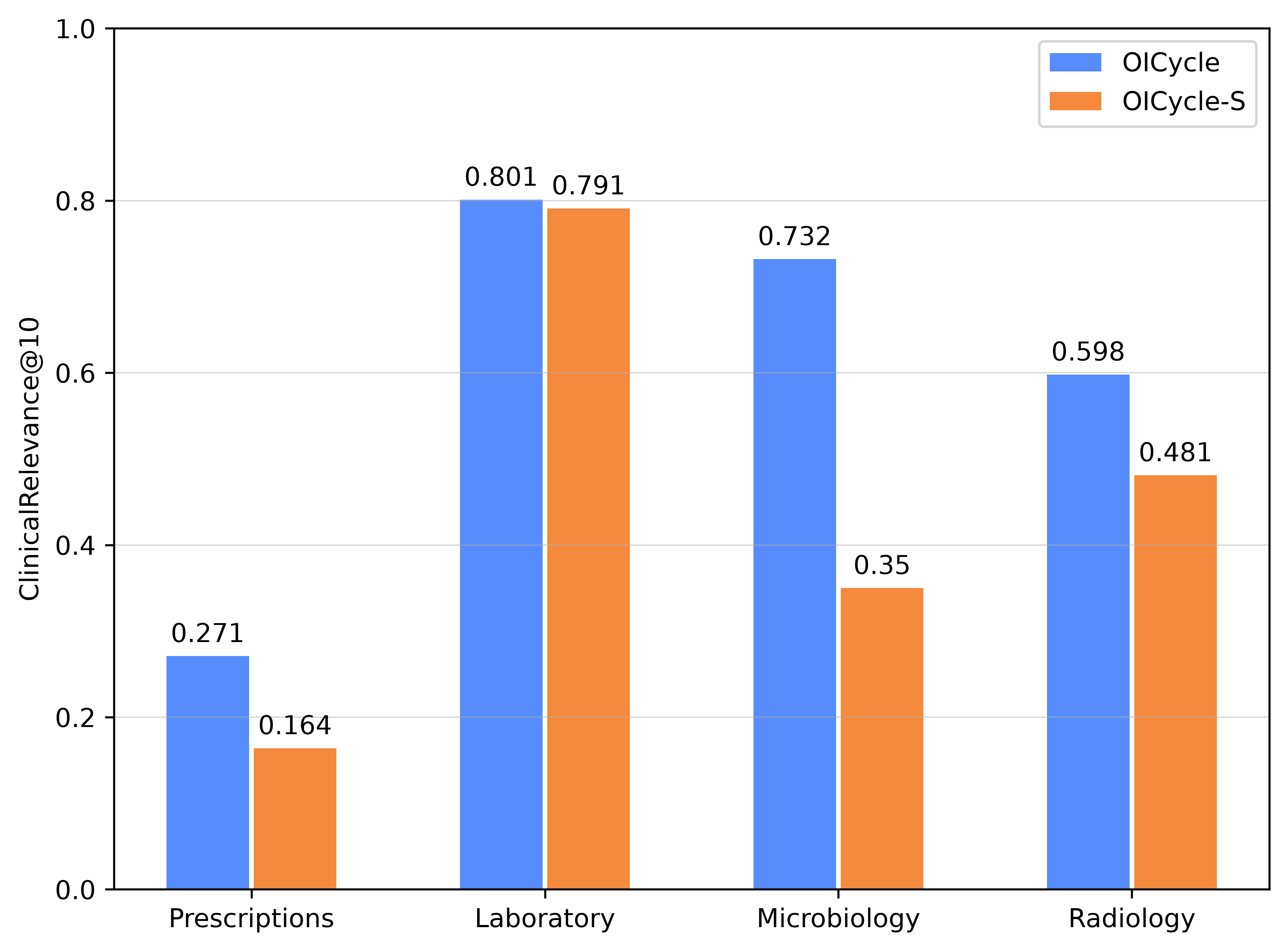}
        \caption{Clinical Relevance@10}
        \label{fig:intervention_cr}
    \end{subfigure}
    \caption{Comparison of intervention-level prediction performance between OICycle \& OICycle-S}
    \label{fig:intervention_performance}
\end{figure}
\subsection{Main Results}
Table \ref{tab:main_result_table} summarizes OICycle and OICycle-S results on all three tasks. OICycle outperforms on Task 1, with a mean AR of 0.34 and a mean CR of 0.60, compared with OICycle-S, which achieves a mean AR of 0.22 and a mean CR of 0.48. For Task 2, OICycle-L leads by 0.19 points with an F1 score of 0.35; for Task 3, it leads by 0.15 points with an F1 score of 0.69.

Figure \ref{fig:intervention_performance} shows intervention-level performance across categories. Laboratory interventions achieve the highest performance, while prescriptions are lowest, with microbiology and radiology in between. This likely reflects the structured, numerical nature of laboratory test data, which supports more accurate predictions. In contrast, prescriptions involve indirect and confounded relationships between orders and outcomes, making prediction more challenging. Microbiology and radiology data include more unstructured or qualitative elements, such as free-text reports, further complicating predictive modeling. These results highlight that input data structure and interpretability are key determinants of predictive accuracy across clinical intervention types.

\section{Analysis \& Discussion}
\begin{figure}[th!]
    \centering
    \begin{subfigure}[b]{0.49\textwidth}
        \centering
        \caption{Clinical Intervention structure}
        \includegraphics[width=\textwidth]{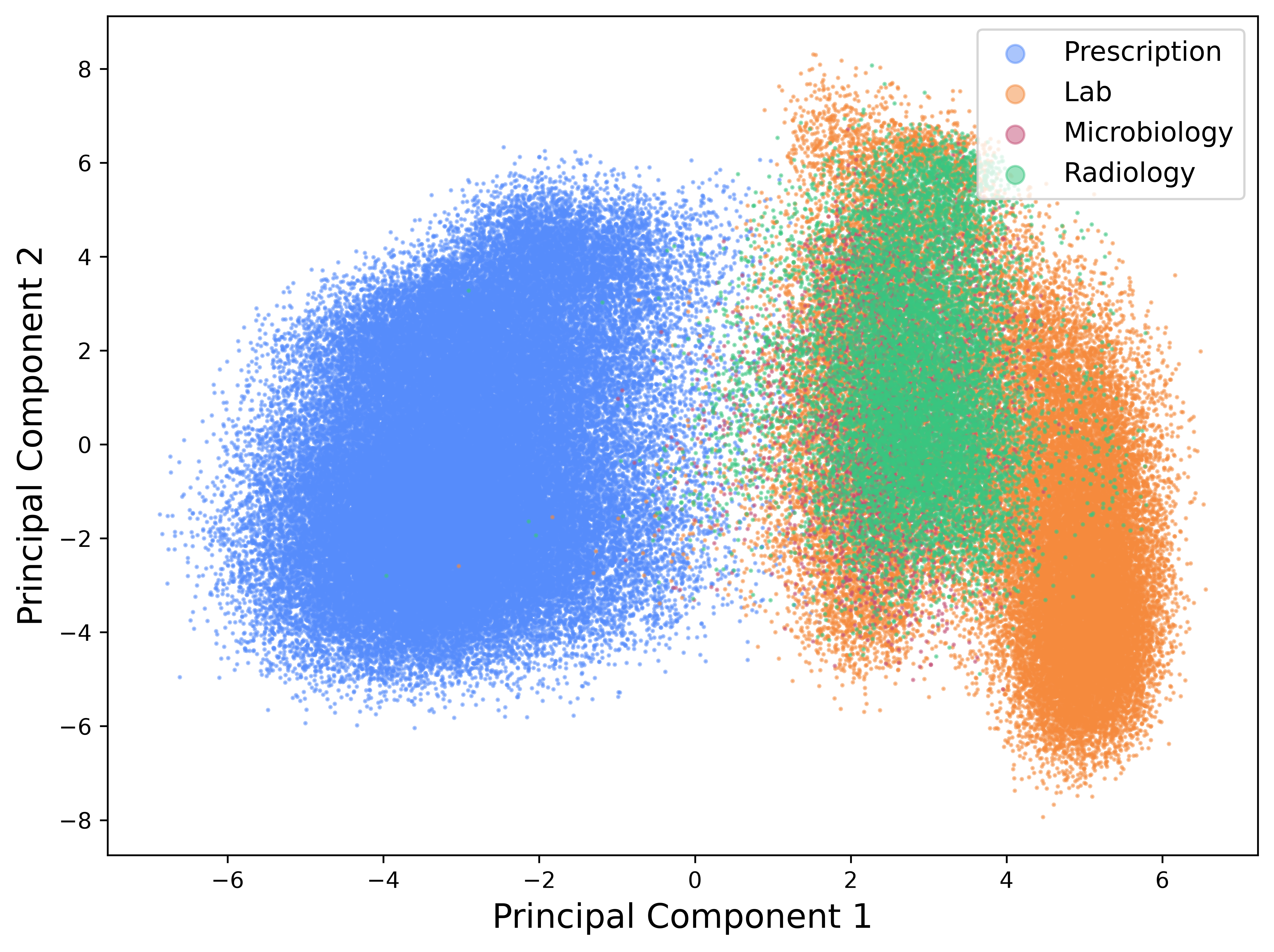}
        \label{fig:panel_a_a}
    \end{subfigure}
    \hfill
    \begin{subfigure}[b]{0.49\textwidth}
        \centering
        \caption{Temporal structure}
        \includegraphics[width=\textwidth]{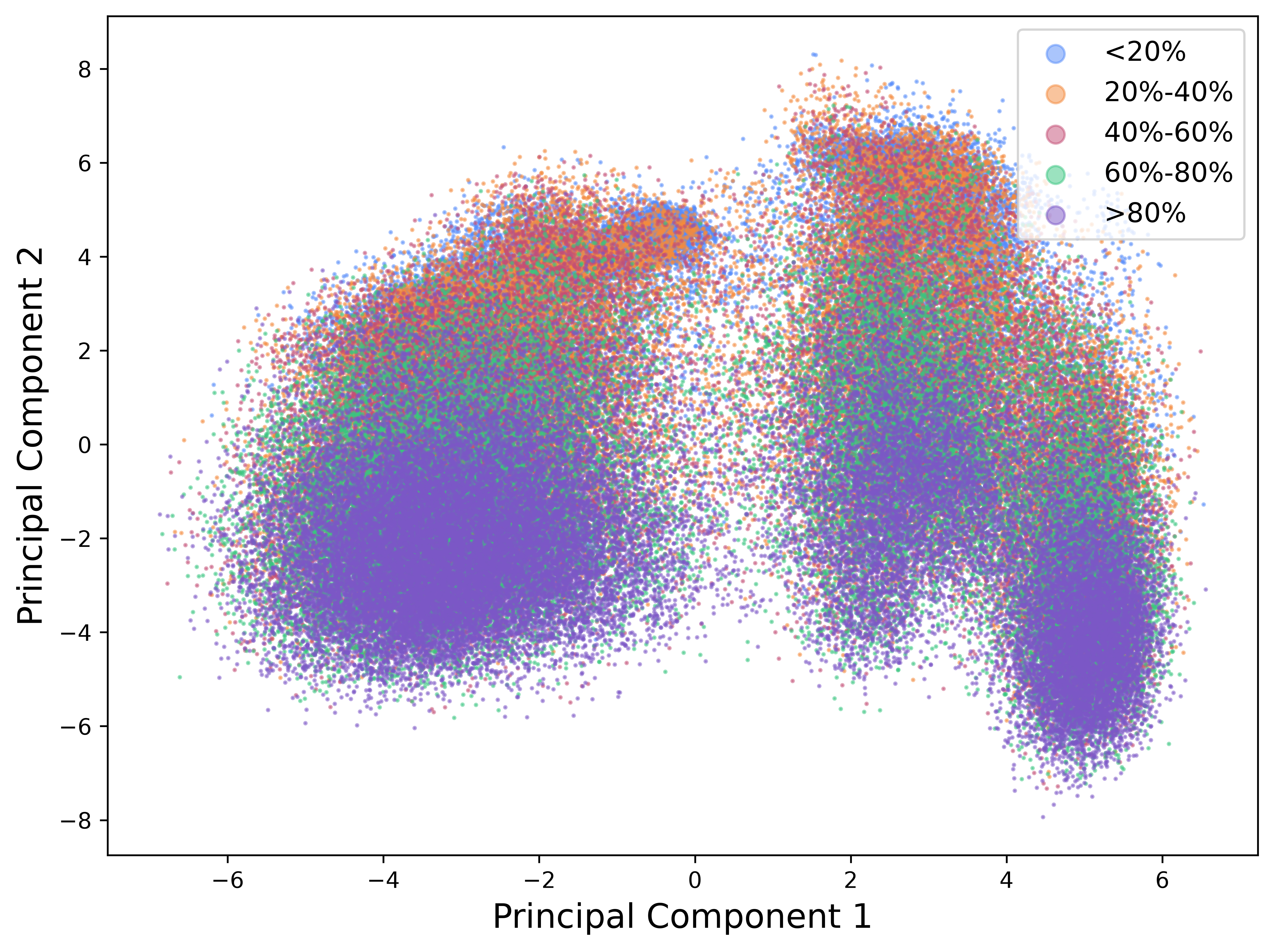}
        \label{fig:panel_a_b}
    \end{subfigure}

    \begin{subfigure}[b]{0.49\textwidth}
        \centering
        \caption{Temporal state difference}
        \includegraphics[width=\textwidth]{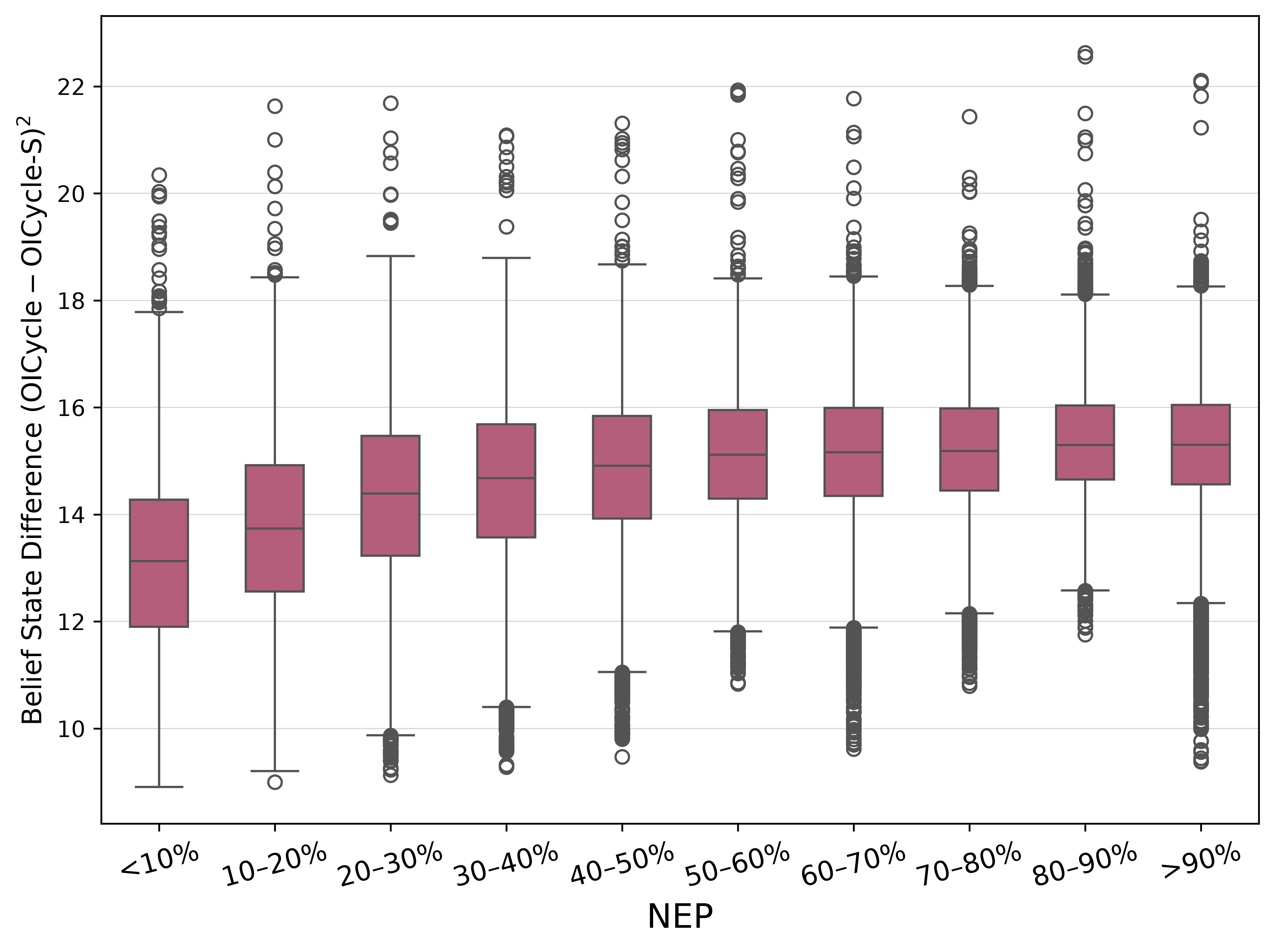}
        \label{fig:panel_b_a}
    \end{subfigure}
    \hfill
    \begin{subfigure}[b]{0.49\textwidth}
        \centering
        \caption{Consecutive state similarity}
        \includegraphics[width=\textwidth]{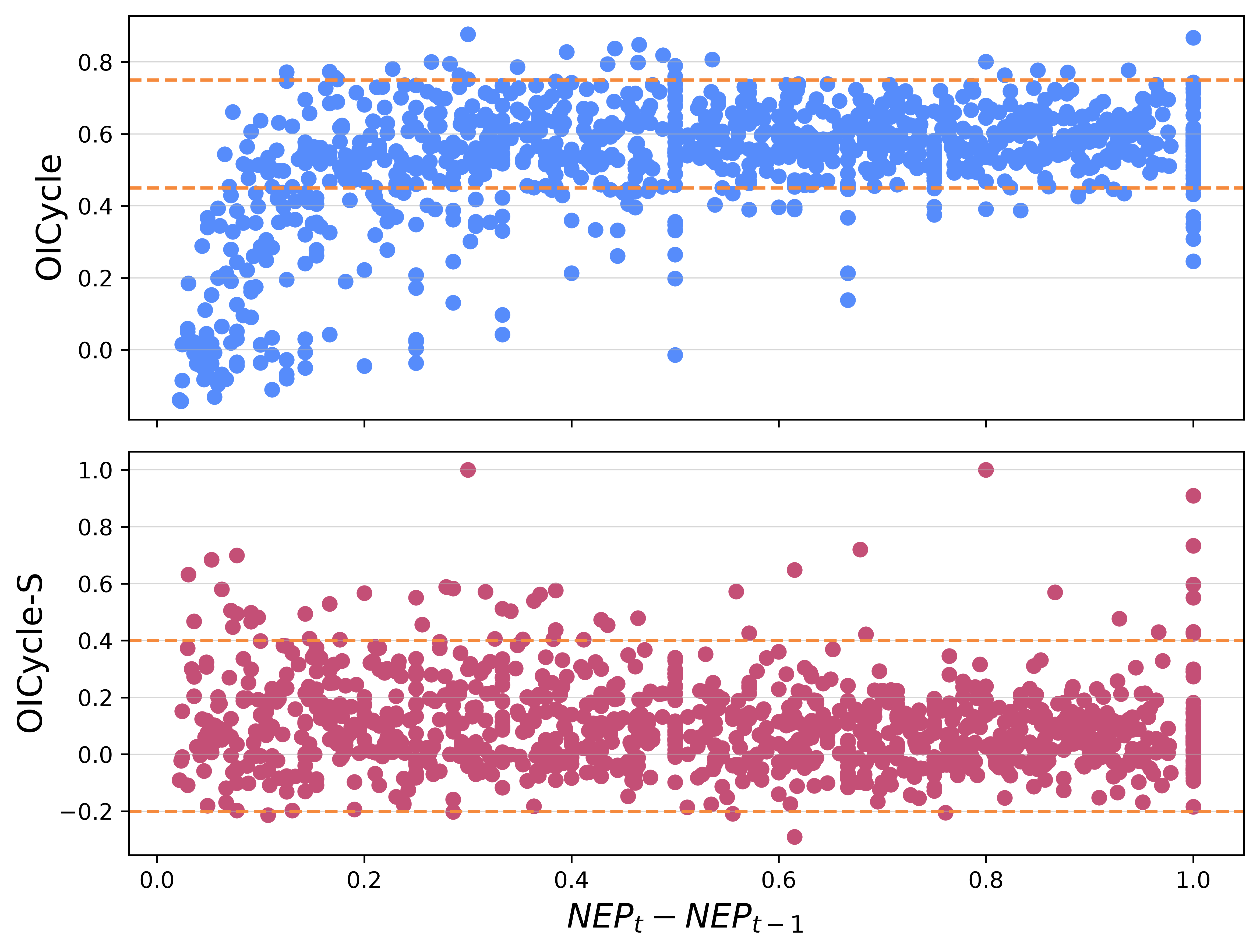}
        \label{fig:panel_b_b}
    \end{subfigure}

    \begin{subfigure}[b]{0.49\textwidth}
        \centering
        \caption{Episode Progression vs Performance}
        \includegraphics[width=\textwidth]{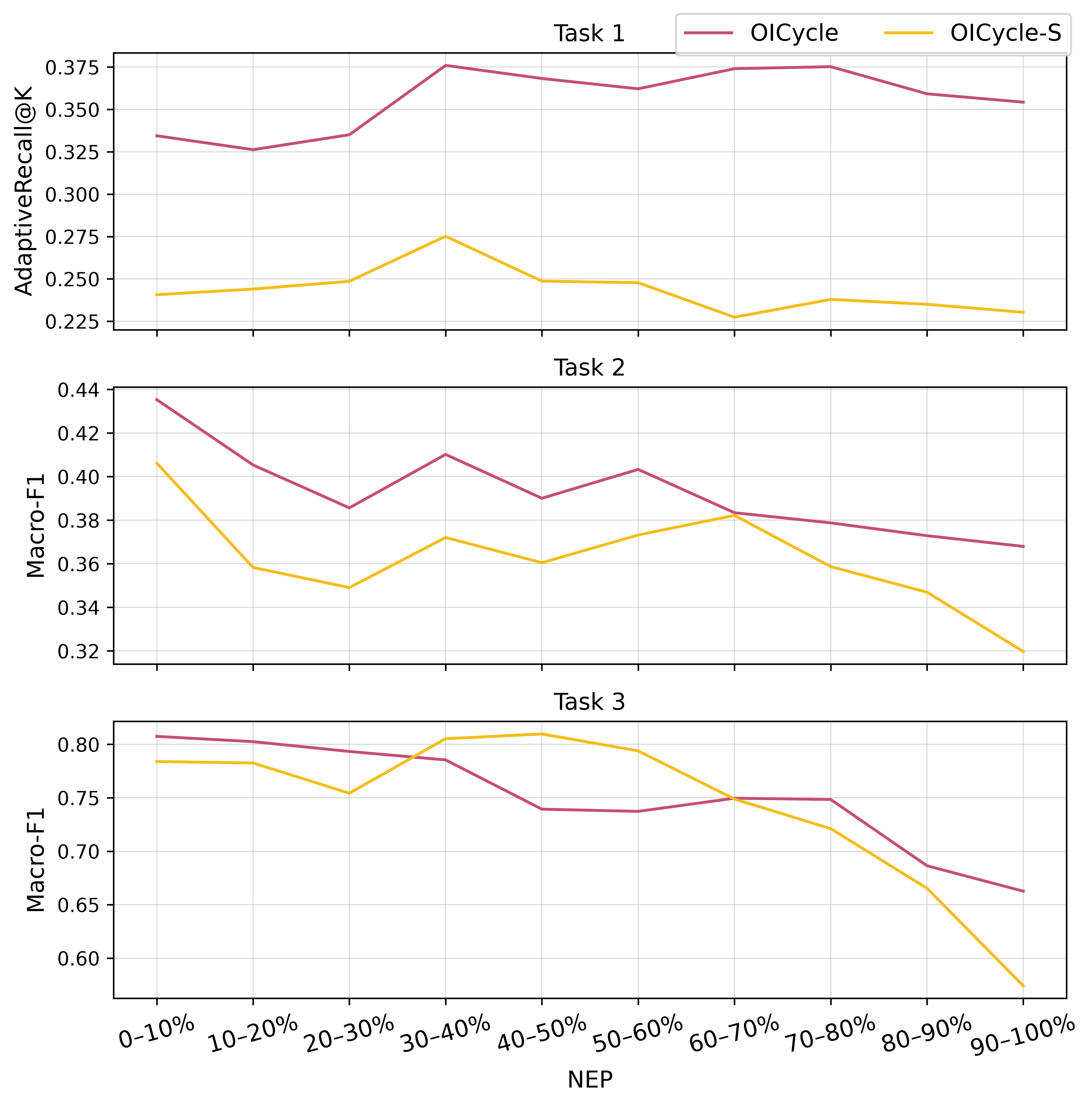}
        \label{fig:panel_c_a}
    \end{subfigure}
    \hfill
    \begin{subfigure}[b]{0.49\textwidth}
        \centering
        \caption{Episode Progression vs Task 1 performance}
        \includegraphics[width=\textwidth]{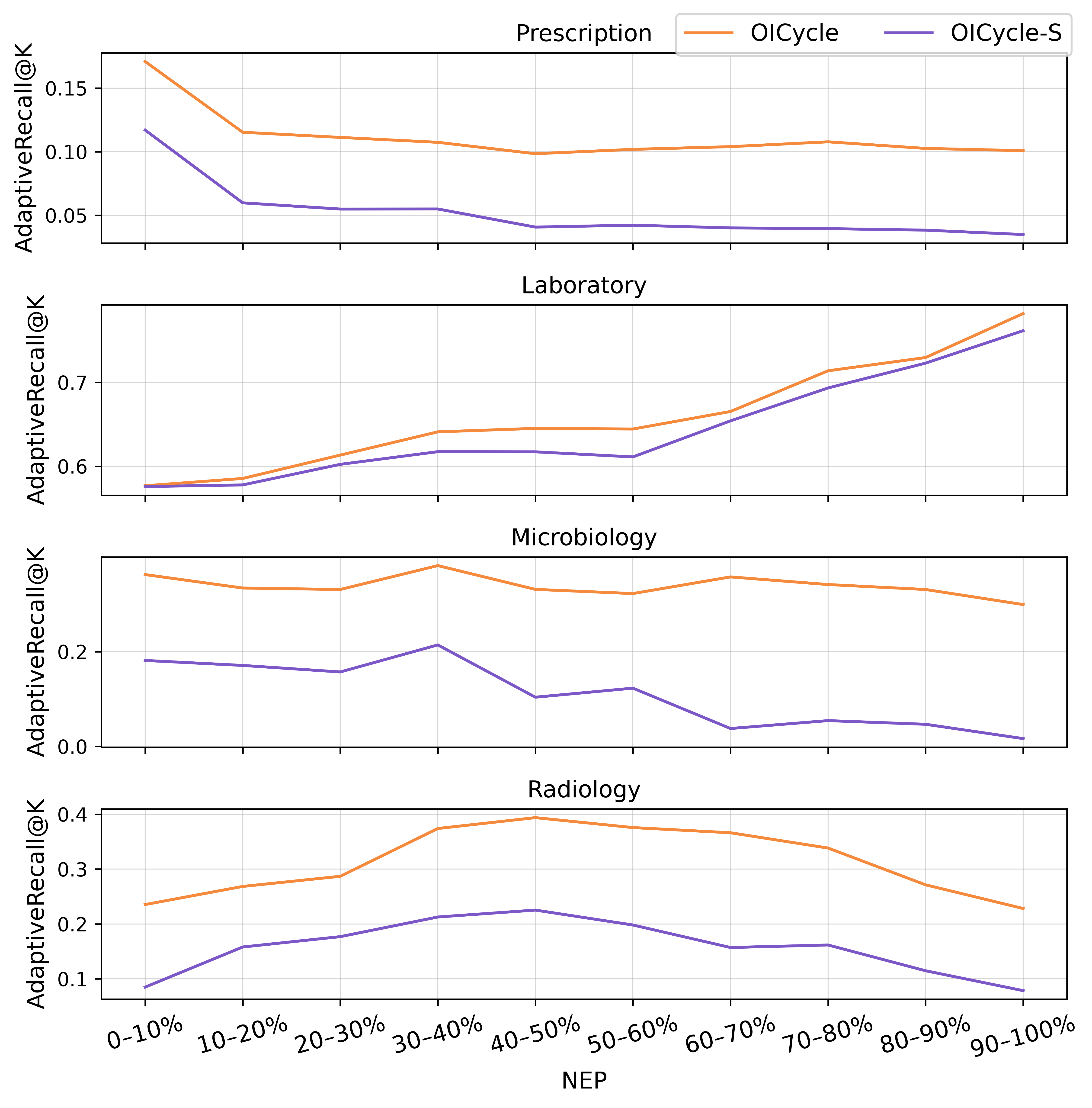}
        \label{fig:panel_c_b}
    \end{subfigure}
    \caption{Figure (a)(b) examines the clinical intervention and temporal structure of the belief state, Fig. (c)(d) examines the evolving nature of the belief state, and Fig. (e)(f) examines the predictive utility of history accumulation in the belief state. Here, NEP means Normalized Episode Progression}
    \label{fig:analysis}
    
\end{figure}
We analyze the learned predictive belief state from three perspectives: representational structure, state dynamics, and the role of history accumulation (Figure \ref{fig:analysis}).

First, we examine what clinically relevant information is encoded in the belief state, including intervention structure and temporal progression (Panel A). Panel A (a) examines the clinical intervention structure in the latent belief state using Principal Component Analysis (PCA). The representations show clear separation by clinical intervention type, particularly between TI and DI interventions. This separation suggests that the belief state distinguishes treatment-related interventions from diagnostic investigations. Panel A (b) examines the temporal structure by grouping representations according to normalized episode progress. The projected representations show differentiation from top to bottom at 0-100\% normalized episode progress, suggesting a structure tied to temporal position in the latent space. 

Second, we examine state dynamics and differences between OICycle and OICycle-S by looking at history dependence and consecutive-state similarity. Panel B(a) examines the relationship between history accumulation and the difference in belief state between OICycle and OICycle-S. As OICycle accumulates more history, the difference between the two models increases, suggesting that OICycle is not just learning a different representation but that this consistently drives differentiation for patients. Panel B(b) examines the cosine similarity between belief states at consecutive episodes for both models. OICycle belief states show a high degree of similarity with previous belief states, reflecting temporal dependency rather than random belief states at each episode. Unlike OICycle, OICycle-S has little similarity, which is trivial, as it does not accumulate history.

Lastly, we examine the relationship between episode progression and performance in Panel C. Panel C(a) \& (b) plot the normalized episode progression against task metrics and intervention metrics. Task 1 metrics and the intervention subtypes are minimally affected by episode progression. On the other hand, task 2 \& task 3 are clearly affected, which can be seen after 60-70\% episode progression, where OICycle and OICycle-S have different slopes for degraded performance compared to early episodes, where their slopes are similar. This suggests that as episodes progress, task2 \& task3 will have a major performance impact, as also seen in the quantitative evaluation. Note that for Task 2 \& 3, macro-F1 is calculated at the episode level, not the dataset level.

In summary, OICycle belief states are not optimized random states but have meaningful structure, relationships with episode progression, and impact due to episode progression. However, several limitations and caveats remain. Because OICycle is trained on observational EHR data, the learned latent dynamics should not be interpreted as uniquely identifying the underlying physiological disease dynamics. Rather, the learned state transition may capture a mixture of patient-related dynamics and clinician behavioral or intervention-application patterns. In addition, the dataset is still smaller than the multimodal scenarios clinicians use for decision-making.  

\section{Conclusion \& Future Work}
In summary, OICycle offers a POMDP-inspired framework for sequential clinical decision-making, updating belief states as interventions and observations unfold. It outperforms OICycle-S, especially for longitudinal outcomes like near-term mortality, and its open dataset establishes a benchmark for future research. Future work includes expanding multimodal data sources (e.g., electronic medication, ECG), clinically validating top predictions, enhancing explainability, and developing tools for clinician training and scenario analysis based on OICycle.

\bibliographystyle{alpha}
\bibliography{iclr}

@article{wornow_ehrshot_nodate,
	title = {{EHRSHOT}: An {EHR} Benchmark for Few-Shot Evaluation of Foundation Models},
	author = {Wornow, Michael and Thapa, Rahul and Steinberg, Ethan and Fries, Jason A and Shah, Nigam H},
	langid = {english},
}

@article{johnson_mimic-iv_2023,
	title = {{MIMIC}-{IV}, a freely accessible electronic health record dataset},
	volume = {10},
	rights = {2023 The Author(s)},
	issn = {2052-4463},
	url = {https://www.nature.com/articles/s41597-022-01899-x},
	doi = {10.1038/s41597-022-01899-x},
	pages = {1},
	number = {1},
	journaltitle = {Scientific Data},
	shortjournal = {Sci Data},
	author = {Johnson, Alistair E. W. and Bulgarelli, Lucas and Shen, Lu and Gayles, Alvin and Shammout, Ayad and Horng, Steven and Pollard, Tom J. and Hao, Sicheng and Moody, Benjamin and Gow, Brian and Lehman, Li-wei H. and Celi, Leo A. and Mark, Roger G.},
	urldate = {2026-08-17},
	date = {2023-01-03},
	langid = {english},
	note = {Publisher: Nature Publishing Group},
}

@article{yeche_hirid-icu-benchmark_nodate,
	title = {{HiRID}-{ICU}-Benchmark — A Comprehensive Machine Learning Benchmark on High-resolution {ICU} Data},
	author = {Yèche, Hugo and Kuznetsova, Rita and Zimmermann, Marc and Hüser, Matthias and Lyu, Xinrui and Faltys, Martin and Rätsch, Gunnar},
	langid = {english},
}

@article{acosta_multimodal_2022,
	title = {Multimodal biomedical {AI}},
	volume = {28},
	rights = {2022 Springer Nature America, Inc.},
	issn = {1546-170X},
	url = {https://www.nature.com/articles/s41591-022-01981-2},
	doi = {10.1038/s41591-022-01981-2},
	pages = {1773--1784},
	number = {9},
	journaltitle = {Nature Medicine},
	shortjournal = {Nat Med},
	author = {Acosta, Julián N. and Falcone, Guido J. and Rajpurkar, Pranav and Topol, Eric J.},
	urldate = {2026-08-20},
	date = {2022-09},
	langid = {english},
	note = {Publisher: Nature Publishing Group},
}

@article{kaelbling_planning_1998,
	title = {Planning and acting in partially observable stochastic domains},
	volume = {101},
	issn = {0004-3702},
	url = {https://www.sciencedirect.com/science/article/pii/S000437029800023X},
	doi = {10.1016/S0004-3702(98)00023-X},
	pages = {99--134},
	number = {1},
	journaltitle = {Artificial Intelligence},
	shortjournal = {Artificial Intelligence},
	author = {Kaelbling, Leslie Pack and Littman, Michael L. and Cassandra, Anthony R.},
	urldate = {2026-09-20},
	date = {1998-05-01},
}

@misc{kingma_auto-encoding_2022,
	title = {Auto-Encoding Variational Bayes},
	url = {http://arxiv.org/abs/1312.6114},
	doi = {10.48550/arXiv.1312.6114},
	number = {{arXiv}:1312.6114},
	publisher = {{arXiv}},
	author = {Kingma, Diederik P. and Welling, Max},
	urldate = {2026-09-20},
	date = {2022-12-10},
	eprinttype = {arxiv},
	eprint = {1312.6114 [stat.ML]},
}

@misc{ling_bioclinical_2023,
	title = {Bio+Clinical {BERT}, {BERT} Base, and {CNN} Performance Comparison for Predicting Drug-Review Satisfaction},
	url = {http://arxiv.org/abs/2308.03782},
	doi = {10.48550/arXiv.2308.03782},
	number = {{arXiv}:2308.03782},
	publisher = {{arXiv}},
	author = {Ling, Yue},
	urldate = {2026-09-21},
	date = {2023-08-02},
	eprinttype = {arxiv},
	eprint = {2308.03782 [cs.CL]},
}

@inproceedings{cho_learning_2014,
	location = {Doha, Qatar},
	title = {Learning Phrase Representations using {RNN} Encoder–Decoder for Statistical Machine Translation},
	url = {https://aclanthology.org/D14-1179/},
	doi = {10.3115/v1/D14-1179},
	eventtitle = {{EMNLP} 2014},
	pages = {1724--1734},
	booktitle = {Proceedings of the 2014 Conference on Empirical Methods in Natural Language Processing ({EMNLP})},
	publisher = {Association for Computational Linguistics},
	author = {Cho, Kyunghyun and van Merriënboer, Bart and Gulcehre, Caglar and Bahdanau, Dzmitry and Bougares, Fethi and Schwenk, Holger and Bengio, Yoshua},
	editor = {Moschitti, Alessandro and Pang, Bo and Daelemans, Walter},
	urldate = {2026-09-21},
	date = {2014-10},
}

@article{najafi_advancements_2026,
	title = {Advancements in reinforcement learning for clinical decision-making in healthcare: a systematic review},
	volume = {4},
	issn = {2731-6955},
	url = {https://doi.org/10.1007/s44248-026-00112-3},
	doi = {10.1007/s44248-026-00112-3},
	shorttitle = {Advancements in reinforcement learning for clinical decision-making in healthcare},
	pages = {13},
	number = {1},
	journaltitle = {Discover Data},
	shortjournal = {Discov Data},
	author = {Najafi, Ali and Farhadi, Amirfarhad and Zamanifar, Azadeh},
	urldate = {2026-09-22},
	date = {2026-07-14},
	langid = {english},
}

@article{li_behrt_2020,
	title = {{BEHRT}: Transformer for Electronic Health Records},
	volume = {10},
	issn = {2045-2322},
	url = {https://pmc.ncbi.nlm.nih.gov/articles/PMC7189231/},
	doi = {10.1038/s41598-020-62922-y},
	shorttitle = {{BEHRT}},
	pages = {7155},
	journaltitle = {Scientific Reports},
	shortjournal = {Sci Rep},
	author = {Li, Yikuan and Rao, Shishir and Solares, José Roberto Ayala and Hassaine, Abdelaali and Ramakrishnan, Rema and Canoy, Dexter and Zhu, Yajie and Rahimi, Kazem and Salimi-Khorshidi, Gholamreza},
	urldate = {2026-09-22},
	date = {2020-04-28},
	pmid = {32346050},
	pmcid = {PMC7189231},
}

@article{rasmy_med-bert_2021,
	title = {Med-{BERT}: pretrained contextualized embeddings on large-scale structured electronic health records for disease prediction},
	volume = {4},
	rights = {2021 The Author(s)},
	issn = {2398-6352},
	url = {https://www.nature.com/articles/s41746-021-00455-y},
	doi = {10.1038/s41746-021-00455-y},
	shorttitle = {Med-{BERT}},
	pages = {86},
	number = {1},
	journaltitle = {npj Digital Medicine},
	shortjournal = {npj Digit. Med.},
	author = {Rasmy, Laila and Xiang, Yang and Xie, Ziqian and Tao, Cui and Zhi, Degui},
	urldate = {2026-09-22},
	date = {2021-05-20},
	langid = {english},
	note = {Publisher: Nature Publishing Group},
}

@article{yu_reinforcement_2021,
	title = {Reinforcement Learning in Healthcare: A Survey},
	volume = {55},
	issn = {0360-0300},
	url = {https://dl.acm.org/doi/10.1145/3477600},
	doi = {10.1145/3477600},
	shorttitle = {Reinforcement Learning in Healthcare},
	pages = {5:1--5:36},
	number = {1},
	journaltitle = {{ACM} Computing Surveys ({CSUR})},
	shortjournal = {{ACM} Comput. Surv.},
	author = {Yu, Chao and Liu, Jiming and Nemati, Shamim and Yin, Guosheng},
	urldate = {2026-09-22},
	date = {2021-11-23},
}

@article{jayaraman_primer_2024,
	title = {A Primer on Reinforcement Learning in Medicine for Clinicians},
	volume = {7},
	issn = {2398-6352},
	url = {https://pmc.ncbi.nlm.nih.gov/articles/PMC11599275/},
	doi = {10.1038/s41746-024-01316-0},
	pages = {337},
	journaltitle = {{NPJ} Digital Medicine},
	shortjournal = {{NPJ} Digit Med},
	author = {Jayaraman, Pushkala and Desman, Jacob and Sabounchi, Moein and Nadkarni, Girish N. and Sakhuja, Ankit},
	urldate = {2026-09-22},
	date = {2024-11-26},
	pmid = {39592855},
	pmcid = {PMC11599275},
}

@misc{huang_clinicalbert_2020,
	title = {{ClinicalBERT}: Modeling Clinical Notes and Predicting Hospital Readmission},
	url = {http://arxiv.org/abs/1904.05342},
	doi = {10.48550/arXiv.1904.05342},
	shorttitle = {{ClinicalBERT}},
	number = {{arXiv}:1904.05342},
	publisher = {{arXiv}},
	author = {Huang, Kexin and Altosaar, Jaan and Ranganath, Rajesh},
	urldate = {2026-09-22},
	date = {2020-11-29},
	eprinttype = {arxiv},
	eprint = {1904.05342 [cs.CL]},
}

@article{lee_biobert_2020,
	title = {{BioBERT}: a pre-trained biomedical language representation model for biomedical text mining},
	volume = {36},
	issn = {1367-4803},
	url = {https://doi.org/10.1093/bioinformatics/btz682},
	doi = {10.1093/bioinformatics/btz682},
	shorttitle = {{BioBERT}},
	pages = {1234--1240},
	number = {4},
	journaltitle = {Bioinformatics},
	shortjournal = {Bioinformatics},
	author = {Lee, Jinhyuk and Yoon, Wonjin and Kim, Sungdong and Kim, Donghyeon and Kim, Sunkyu and So, Chan Ho and Kang, Jaewoo},
	urldate = {2026-09-22},
	date = {2020-02-01},
}

@inproceedings{choi_gram_2017,
	location = {New York, {NY}, {USA}},
	title = {{GRAM}: Graph-based Attention Model for Healthcare Representation Learning},
	isbn = {978-1-4503-4887-4},
	url = {https://dl.acm.org/doi/10.1145/3097983.3098126},
	doi = {10.1145/3097983.3098126},
	series = {{KDD} '17},
	shorttitle = {{GRAM}},
	pages = {787--795},
	booktitle = {Proceedings of the 23rd {ACM} {SIGKDD} International Conference on Knowledge Discovery and Data Mining},
	publisher = {Association for Computing Machinery},
	author = {Choi, Edward and Bahadori, Mohammad Taha and Song, Le and Stewart, Walter F. and Sun, Jimeng},
	urldate = {2026-09-22},
	date = {2017-08-04},
}

@misc{peng_transfer_2019,
	title = {Transfer Learning in Biomedical Natural Language Processing: An Evaluation of {BERT} and {ELMo} on Ten Benchmarking Datasets},
	url = {http://arxiv.org/abs/1906.05474},
	doi = {10.48550/arXiv.1906.05474},
	shorttitle = {Transfer Learning in Biomedical Natural Language Processing},
	number = {{arXiv}:1906.05474},
	publisher = {{arXiv}},
	author = {Peng, Yifan and Yan, Shankai and Lu, Zhiyong},
	urldate = {2026-09-22},
	date = {2019-06-18},
	eprinttype = {arxiv},
	eprint = {1906.05474 [cs.CL]},
}

@inproceedings{devlin_bert_2019,
	location = {Minneapolis, Minnesota},
	title = {{BERT}: Pre-training of Deep Bidirectional Transformers for Language Understanding},
	url = {https://aclanthology.org/N19-1423/},
	doi = {10.18653/v1/N19-1423},
	shorttitle = {{BERT}},
	eventtitle = {{NAACL}-{HLT} 2019},
	pages = {4171--4186},
	booktitle = {Proceedings of the 2019 Conference of the North American Chapter of the Association for Computational Linguistics: Human Language Technologies, Volume 1 (Long and Short Papers)},
	publisher = {Association for Computational Linguistics},
	author = {Devlin, Jacob and Chang, Ming-Wei and Lee, Kenton and Toutanova, Kristina},
	editor = {Burstein, Jill and Doran, Christy and Solorio, Thamar},
	urldate = {2026-09-22},
	date = {2019-06},
}

@article{pham_predicting_2017,
	title = {Predicting healthcare trajectories from medical records: A deep learning approach},
	volume = {69},
	issn = {1532-0464},
	url = {https://www.sciencedirect.com/science/article/pii/S1532046417300710},
	doi = {10.1016/j.jbi.2017.04.001},
	shorttitle = {Predicting healthcare trajectories from medical records},
	pages = {218--229},
	journaltitle = {Journal of Biomedical Informatics},
	shortjournal = {Journal of Biomedical Informatics},
	author = {Pham, Trang and Tran, Truyen and Phung, Dinh and Venkatesh, Svetha},
	urldate = {2026-09-22},
	date = {2017-05-01},
}

@misc{li_optimizing_2019,
	title = {Optimizing Sequential Medical Treatments with Auto-Encoding Heuristic Search in {POMDPs}},
	url = {http://arxiv.org/abs/1905.07465},
	doi = {10.48550/arXiv.1905.07465},
	number = {{arXiv}:1905.07465},
	publisher = {{arXiv}},
	author = {Li, Luchen and Komorowski, Matthieu and Faisal, Aldo A.},
	urldate = {2026-09-22},
	date = {2019-05-17},
	eprinttype = {arxiv},
	eprint = {1905.07465 [cs.AI]},
}

@article{hauskrecht_planning_2000,
	title = {Planning treatment of ischemic heart disease with partially observable Markov decision processes},
	volume = {18},
	issn = {0933-3657},
	url = {https://www.sciencedirect.com/science/article/pii/S0933365799000421},
	doi = {10.1016/S0933-3657(99)00042-1},
	pages = {221--244},
	number = {3},
	journaltitle = {Artificial Intelligence in Medicine},
	shortjournal = {Artificial Intelligence in Medicine},
	author = {Hauskrecht, Milos and Fraser, Hamish},
	urldate = {2026-09-22},
	date = {2000-03-01},
}

@article{tsoukalas_data_2015,
	title = {From Data to Optimal Decision Making: A Data-Driven, Probabilistic Machine Learning Approach to Decision Support for Patients With Sepsis},
	volume = {3},
	issn = {2291-9694},
	url = {https://pmc.ncbi.nlm.nih.gov/articles/PMC4376114/},
	doi = {10.2196/medinform.3445},
	shorttitle = {From Data to Optimal Decision Making},
	pages = {e11},
	number = {1},
	journaltitle = {{JMIR} Medical Informatics},
	shortjournal = {{JMIR} Med Inform},
	author = {Tsoukalas, Athanasios and Albertson, Timothy and Tagkopoulos, Ilias},
	urldate = {2026-09-22},
	date = {2015-02-24},
	pmid = {25710907},
	pmcid = {PMC4376114},
}

@inproceedings{chen_flow-based_2022,
	title = {Flow-based Recurrent Belief State Learning for {POMDPs}},
	url = {https://proceedings.mlr.press/v162/chen22q.html},
	eventtitle = {International Conference on Machine Learning},
	pages = {3444--3468},
	booktitle = {Proceedings of the 39th International Conference on Machine Learning},
	publisher = {{PMLR}},
	author = {Chen, Xiaoyu and Mu, Yao Mark and Luo, Ping and Li, Shengbo and Chen, Jianyu},
	urldate = {2026-09-22},
	date = {2022-06-28},
	langid = {english},
	note = {{ISSN}: 2640-3498},
}

@article{li_beyond_2026,
	title = {Beyond static risk scores: dynamic world models simulating patient-specific trajectories to inform preoperative risk mitigation strategies},
	volume = {13},
	issn = {2296-858X},
	url = {https://www.frontiersin.org/journals/medicine/articles/10.3389/fmed.2026.1913206/full},
	doi = {10.3389/fmed.2026.1913206},
	shorttitle = {Beyond static risk scores},
	journaltitle = {Frontiers in Medicine},
	shortjournal = {Front. Med.},
	author = {Li, Jiawen and Fang, Xuchao and Yu, Ping and Jiang, Lanfang},
	urldate = {2026-09-22},
	date = {2026-09-17},
	note = {Publisher: Frontiers},
}

@article{allenbrand_decide-lab_2026,
	title = {{DECIDE}-Lab: A Value-of-Information and {POMDP} Framework for Diagnostic Laboratory Test Selection},
	volume = {16},
	issn = {2075-4418},
	url = {https://pmc.ncbi.nlm.nih.gov/articles/PMC13464772/},
	doi = {10.3390/diagnostics16152356},
	shorttitle = {{DECIDE}-Lab},
	pages = {2356},
	number = {15},
	journaltitle = {Diagnostics},
	shortjournal = {Diagnostics (Basel)},
	author = {Allenbrand, Corban},
	urldate = {2026-09-22},
	date = {2026-07-27},
	pmid = {42587593},
	pmcid = {PMC13464772},
}

@misc{krishnan_deep_2015,
	title = {Deep Kalman Filters},
	url = {http://arxiv.org/abs/1511.05121},
	doi = {10.48550/arXiv.1511.05121},
	number = {{arXiv}:1511.05121},
	publisher = {{arXiv}},
	author = {Krishnan, Rahul G. and Shalit, Uri and Sontag, David},
	urldate = {2026-09-24},
	date = {2015-11-25},
	eprinttype = {arxiv},
	eprint = {1511.05121 [stat.ML]},
}

@inproceedings{futoma_popcorn_2020,
	title = {{POPCORN}: Partially Observed Prediction Constrained Reinforcement Learning},
	url = {https://proceedings.mlr.press/v108/futoma20a.html},
	shorttitle = {{POPCORN}},
	eventtitle = {International Conference on Artificial Intelligence and Statistics},
	pages = {3578--3588},
	booktitle = {Proceedings of the Twenty Third International Conference on Artificial Intelligence and Statistics},
	publisher = {{PMLR}},
	author = {Futoma, Joseph and Hughes, Michael and Doshi-Velez, Finale},
	urldate = {2026-09-24},
	date = {2020-06-03},
	langid = {english},
	note = {{ISSN}: 2640-3498},
}

@inproceedings{yang_pyhealth_2023,
	location = {Long Beach {CA} {USA}},
	title = {{PyHealth}: A Deep Learning Toolkit for Healthcare Applications},
	isbn = {979-8-4007-0103-0},
	url = {https://dl.acm.org/doi/10.1145/3580305.3599178},
	doi = {10.1145/3580305.3599178},
	shorttitle = {{PyHealth}},
	eventtitle = {{KDD} '23: The 29th {ACM} {SIGKDD} Conference on Knowledge Discovery and Data Mining},
	pages = {5788--5789},
	booktitle = {Proceedings of the 29th {ACM} {SIGKDD} Conference on Knowledge Discovery and Data Mining},
	publisher = {{ACM}},
	author = {Yang, Chaoqi and Wu, Zhenbang and Jiang, Patrick and Lin, Zhen and Gao, Junyi and Danek, Benjamin P. and Sun, Jimeng},
	urldate = {2026-09-24},
	date = {2023-08-06},
	langid = {english},
}

@inproceedings{wang_mimic-extract_2020,
	location = {Toronto Ontario Canada},
	title = {{MIMIC}-Extract: a data extraction, preprocessing, and representation pipeline for {MIMIC}-{III}},
	isbn = {978-1-4503-7046-2},
	url = {https://dl.acm.org/doi/10.1145/3368555.3384469},
	doi = {10.1145/3368555.3384469},
	shorttitle = {{MIMIC}-Extract},
	eventtitle = {{ACM} {CHIL} '20: {ACM} Conference on Health, Inference, and Learning},
	pages = {222--235},
	booktitle = {Proceedings of the {ACM} Conference on Health, Inference, and Learning},
	publisher = {{ACM}},
	author = {Wang, Shirly and {McDermott}, Matthew B. A. and Chauhan, Geeticka and Ghassemi, Marzyeh and Hughes, Michael C. and Naumann, Tristan},
	urldate = {2026-09-24},
	date = {2020-04-02},
	langid = {english},
}

@article{pollard_physionet_2026, title = {{PhysioNet} as a global platform for biomedical research}, volume = {1}, issn = {3005-0693}, url = {https://doi.org/10.1038/s44360-026-00096-z}, doi = {10.1038/s44360-026-00096-z}, number = {8}, journal = {Nature Health}, author = {Pollard, Tom and Moody, Benjamin E. and Lehman, Li-wei H. and Gow, Brian J. and Fernandes, Chrystinne and Xie, Chen and Johnson, Alistair and Mark, Roger G. and Heldt, Thomas}, month = aug, year = {2026}, pages = {792--795},}

@misc{johnson_mimic-cxr-jpg_2019,
	title = {{MIMIC}-{CXR}-{JPG}, a large publicly available database of labeled chest radiographs},
	url = {http://arxiv.org/abs/1901.07042},
	doi = {10.48550/arXiv.1901.07042},
	number = {{arXiv}:1901.07042},
	publisher = {{arXiv}},
	author = {Johnson, Alistair E. W. and Pollard, Tom J. and Greenbaum, Nathaniel R. and Lungren, Matthew P. and Deng, Chih-ying and Peng, Yifan and Lu, Zhiyong and Mark, Roger G. and Berkowitz, Seth J. and Horng, Steven},
	urldate = {2026-09-26},
	date = {2019-11-14},
	eprinttype = {arxiv},
	eprint = {1901.07042 [cs.CV]},
}

@article{PhysioNet-mimic-iv-note-2.2,
  author = {Johnson, Alistair and Pollard, Tom and Horng, Steven and Celi, Leo Anthony and Mark, Roger},
  title = {{MIMIC-IV-Note: Deidentified free-text clinical notes}},
  journal = {{PhysioNet}},
  year = {2023},
  month = jan,
  note = {Version 2.2},
  doi = {10.13026/1n74-ne17},
  url = {https://doi.org/10.13026/1n74-ne17}
}

\end{document}